\documentclass[letterpaper]{article} 
\usepackage{colm2024_conference}

\usepackage{microtype}
\usepackage{hyperref}
\usepackage{url}
\usepackage{booktabs}
\usepackage[dvips]{graphicx}
\usepackage{graphicx}
\usepackage{natbib}
\usepackage{listings}
\definecolor{darkblue}{rgb}{0, 0, 0.5}
\hypersetup{colorlinks=true, citecolor=darkblue, linkcolor=darkblue, urlcolor=darkblue}

\title{Probing the Robustness of Theory of Mind \\ in Large Language Models}

\author{Christian Nickel, Laura Schrewe, Lucie Flek 
\\
Department of Computer Science / \\
Bonn-Aachen International Center for IT\\
University of Bonn\\
\texttt{cnickel@uni-bonn.de, schrewe@uni-bonn.de, lflek@bit.uni-bonn.de} \\
}

\colmfinalcopy 
\begin{document}

\maketitle

\begin{abstract}
    With the success of ChatGPT and other similarly sized SotA LLMs, claims of emergent human like social reasoning capabilities, especially Theory of Mind (ToM), in these models have appeared in the scientific literature. On the one hand those ToM-capabilities have been successfully tested using tasks styled similar to those used in psychology \citep{kosinskiTheoryMindMight2023}. On the other hand, follow up studies showed that those capabilities vanished when the tasks were slightly altered \citep{ullmanLargeLanguageModels2023}.
    In this work we introduce a novel dataset of 68 tasks for probing ToM in LLMs, including potentially challenging variations which are assigned to 10 complexity classes. This way it is providing novel insights into the challenges LLMs face with those task variations. We evaluate the ToM performance of four SotA open source LLMs on our dataset and the dataset introduced by \citet{kosinskiTheoryMindMight2023}.
    The overall low goal accuracy across all evaluated models indicates only a limited degree of ToM capabilities.
    The LLMs' performance on simple complexity class tasks from both datasets are similar. Whereas we find a consistent tendency in all tested LLMs to perform poorly on tasks that require the realization that an agent has knowledge of automatic state changes in its environment, even when those are spelled out to the model. For task complications that change the relationship between objects by replacing prepositions, we notice a performance drop in all models, with the strongest impact on the mixture-of-experts model.  
    With our dataset of tasks grouped by complexity we offer directions for further research on how to stabilize and advance ToM capabilities in LLM.
\end{abstract}

\section{Introduction}
Theory of Mind (ToM) -  the ability to track the concealed mental states of others, encompassing knowledge, intentions, beliefs, and desires - is considered a facet of social intelligence \cite{heyes2014cultural, zhang2012perspective, blatt2010does,swim2015portraying}. It might help with many applications, for instance programming or chatbot assistance. Whether, as claimed in \citet{kosinskiTheoryMindMight2023} ToM emerged without specific training, just through increasing the size of the model and training data is an open question.  Large language models (LLMs) seem to exhibit ToM capabilities when evaluated on simple tasks commonly used in psychology, namely unexpected transfer or unexpected content tasks \citep{kosinskiTheoryMindMight2023}. But did the model really learn how other agent's minds work and what they think or just linguistic patterns present in standard ToM tasks? \citet{ullmanLargeLanguageModels2023} suggests that the LLMs do not exhibit real ToM capabilities, since they vanished when prompted with slight variations of the original tasks. While \citet{ullmanLargeLanguageModels2023} introduces certain categories of complications that are challenging for LLMs, besides a few examples the authors do not publicly provide a large dataset that would allow more systematical research.

The main contribution of this work is the creation of a novel ToM benchmarking dataset consisting of manually crafted unexpected content and unexpected transfer tasks based on 10 complexity classes. Furthermore we evaluate our new dataset on four SotA LLMs.
Knowing which complications are especially challenging to the SOTA LLMs might help understand how the models' ToM capabilities work internally and might yield directions of research onto how this capabilities can be improved.

\subsection{Related Work}
Several ToM studies and datasets have been released since the original work of \citet{kosinskiTheoryMindMight2023} and \citet{ullmanLargeLanguageModels2023}, yet show notable differences from our approach. \citet{sartoriLanguageModelsPsychological2023} emphasize the potential usefulness of the investigation of LLM capabilities, like ToM, or their errors on related tasks for research into human cognition and biases.
The newly published FANToM dataset and paper \citep{kimFANToMBenchmarkStresstesting2023} focus on dynamic social interactions.
Another recent work, "ToMBench" \citep{chenToMBenchBenchmarkingTheory2024} is a dataset of multiple-choice cognition tasks, evaluated by the authors on several LLMs, leading them to the statement that "even the most advanced LLMs like GPT-4 lag behind human performance by over 10\% points, indicating that LLMs have not achieved a human-level theory of mind yet". Literature on the basics on how models might not really learn specific abilities, like ToM for instance, but much rather "only" linguistic patterns present in the tasks, possibly through "contaminated" training data, resulting in a "stochastic parrot" is \citep{benderDangersStochasticParrots2021a}.
Another critical acount on the ToM abilities can be found in \citep{shapiraCleverHansNeural2023}, in which the authors conduct experiments using 6 tasks probing into different aspects of ToM.

\section{Methodology}
\subsection{Overview}
In order to measure the Theory of Mind (ToM) performance of Large Language Models on basic ToM tasks as well as variations derived from the these basic tasks, we first manually create a novel dataset. Each task variation is assigned one out of 10 complexity classes pertaining to similar kinds of challenges we introduced in said variation. Of course the correct solutions to the tasks are also included. Besides the data necessary for the task at hand the dataset also already entails the belief, which the protagonist of the tasks holds after each sentence. This might be of interest for further research into Chain-of-Thought (CoT) -reasoning and -faithfulness in the context of ToM. In the next step we prepend instructions, meant to make the LLM output more compatible with machine evaluation, to the task and administer the resulting prompt to four State-of-the-Art (SotA) LLMs. We apply prompt tuning techniques as described by \citet{bsharatPrincipledInstructionsAre2024}. The full promt can be found in the appendix \ref{appendix:full_prompt}. Nevertheless the actual LLM output can be complex and we first have to extract the machine-evaluable "final-answer" with our extraction function. Then we are able to determine which task has been answered correctly. We report the overall- as well as the per complexity-class-performance of each model. We investigate which alterations of the tasks might be the most challenging to the models.

To understand our dataset it is easiest to imagine a stage play. Each play takes place on its own stage and is comprised of one or several stage settings respectively sceneries.
For every scene taking place in said scenery there might be several props placed on that stage. Which props are placed and their positions might differ from scene to scene.
Now imagine the stage play has a very experimental approach. In order to make the play interactive and captivating after each scene the audience is asked questions about what just happened, what was where on stage and what the protagonists were thinking. This can be hard as objects containing others might be opaque and incorrectly labelled (ToM unexpected contents task) or transfers of objects might have taken place without the protagonist being present (ToM unexpected transfer tasks).
Questions about the protagonists believes are Theory of Mind questions, whereas the others are checking a general understanding of the scene. They might serve as sanity checks whether the member of the audience truly understands both, the believes of the protagonists and the real world states of the scene or is just giving the "less" obvious answer to weirdly easy questions.
This mental model is the inspiration for the structure of our dataset.

\subsubsection{Dataset Creation and Outline}
\begin{figure}[h]
\begin{center}
\includegraphics[width=1\linewidth]{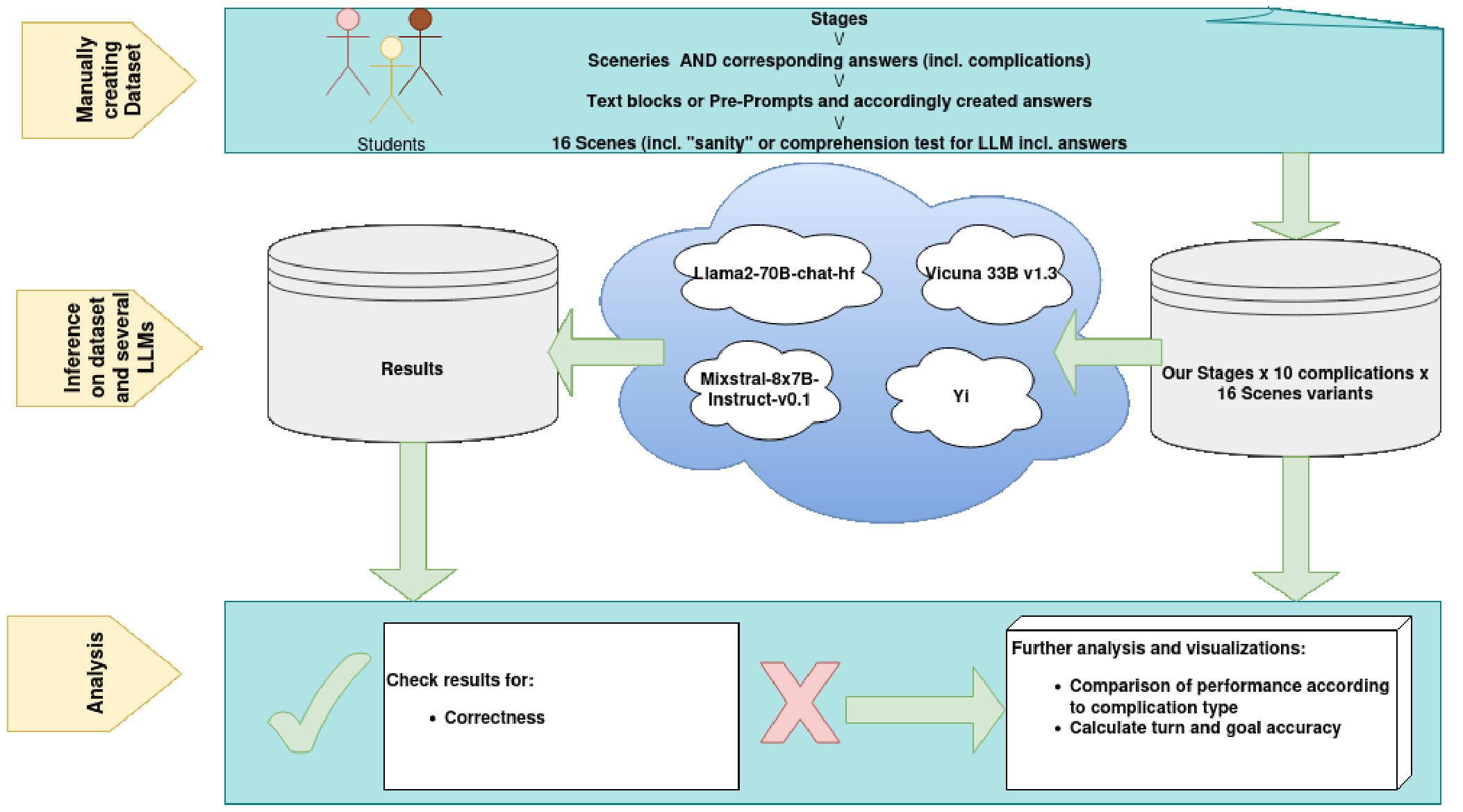}
\end{center}
\caption{Dataset creation and evaluation pipeline}
\label{fig:pipeline_diagram}
\end{figure}

We create seven different stage plays (or short stages).
The basic idea is that within the same stage several sceneries (tasks) and scenes (activities or arrangement of objects) can take place. About each each scene we might ask ToM or general understanding or spatial reasoning questions
(figure \ref{fig:pipeline_diagram}
). 
For each one of the 10 complexity categories we defined beforehand, we manually alter the text in the way characteristic to the respective complexity class. For example, for the complexity class named "transparent container", we take the initial stage text and replace every mention of "intransparent paper bag" with "transparent paper bag" and adapt other parts of the text where necessary. Each of these "complications" of a stage, including the unaltered version, is called a scenery.
Note that for some stages we noticed that some complexity categories are not applicable because the alterations would not make sense in the context of the plot. Therefore the dataset consists of 68 sceneries in total (and not 7 x 11 sceneries).
We generate 16 sub-tasks, called scenes. 

The creation of the scenes follows the procedure by \citet{kosinskiTheoryMindMight2023}. We apply object swaps, add a true belief version, a version where the protagonist gets informed about the true world state and a version where the true world state in visible. Additionally to asking for the protagonists belief, versions that ask for the true world state are introduced, to test the LLMs understanding of the scene.
As a result we obtain 68 x 16 = 1088 scenes with their corresponding correct solutions.
Regarding formal requirement on the texts to avoid unintended hints for the LLM we followed the principles used in the instruction given to research assistants\footnote{\url{https://osf.io/csdhb/wiki/Instructions\%20for\%20RAs/}} by Kosinsky. For instance we made sure, that the key words appear exactly the same number of times to prevent biasing the LLMs' output probabilities.
Out of our seven stages, four represent unexpected content tasks and three are unexpected transfer tasks.

\subsubsection{Complexity Classes}
Additionally to the five complexity classes introduced in \citet{ullmanLargeLanguageModels2023}, we add the following five new complexity classes. Examples for each class can be found in the appendix \ref{appendix:complexity_class_examples}.




    \paragraph{automatic change knowledge}
    To tell the correct world state at the end of the scenery it is required to understand an automatic process, that is independent of the protagonists actions. The process is described in the text.
    
    Example:
    \texttt{"Charlie buys a non-transparent box of green mangos. There are no red mangos in the box. 
    Green mangos ripen, change the color and become red mangos after a few days. Charlie does not know this.[...]"}

    \paragraph{add unrelated information}
    We add additional facts to confuse the model. This facts might be related to the persons or objects in the story, but should be irrelevant for the world state or protagonist belief we will ask for.

    \paragraph{induction from baseline}
    The belief of the protagonist is based on a logical induction. Based on the experience that every time the protagonist observed something a specific fact is true, the person concludes it also holds in the given situation.
    
    \paragraph{untrustworthy testimony}
    This complication is similar to the idea of informed protagonist, but here the protagonist does not trust the other persons' testimony.

    \paragraph{conclusion from sentiment}
    The information is only given indirectly via the sentiment of the protagonist regarding the information.

\subsection{Models and Inference}
We evaluate our dataset on four open source transformer-based state-of-the-art models Llama-2-70-b-chat-hf \citep{touvronLlamaOpenFoundation2023} (70B parameters), Vicuna-33b-v1.3 \citep{VicunaOpenSourceChatbot} (33B parameters),  Yi-34B-Chat \citep{aiYiOpenFoundation2024} (34B parameters) and Mixtral-8x7B-Instruct-v0.1 \citep{aiMixtralExperts2023}, which is a mixture of experts (MoE) model consisting of 8 models with 7B parameters each.

We use a temperature of 1.0 and set the number of maximum output tokens to two times the number of tokens in the respective model input to scale it dynamically based on the input length.

\subsection{Evaluation Metrics}

Using two recursive functions we first identify the words that constitute said answer and then evaluate it against the manually crafted solution given in the dataset. This results in a boolean value for each answer, namely whether it is "True" (i.e. correct) or "False" (i.e. incorrect).
To account for variation in the LLM output like different spelling or spacing we transform the extracted answer to a standardized form before the correctness check. We relax the correctness condition from exact string matching to string inclusion, while we also ensure the wrong solution is not present in the LLMs' answer.
We call the rate of correctly answered scenes turn accuracy.
To get a more meaningful metric for the performance than just the correctness of individual sceneries (turn accuracy), we introduce the notion of goal accuracy. We call a scenery answered goal accurate if all scenes belonging to the scenery have been answered correctly. Thus we ensure a comprehensive understanding of the given situation. We also calculate these measures for the complexity classes in order to find out whether certain categories are easier or more challenging to the models tested.


\section{Results}
\begin{figure}[t]
\begin{center}
\includegraphics[width=1\linewidth]{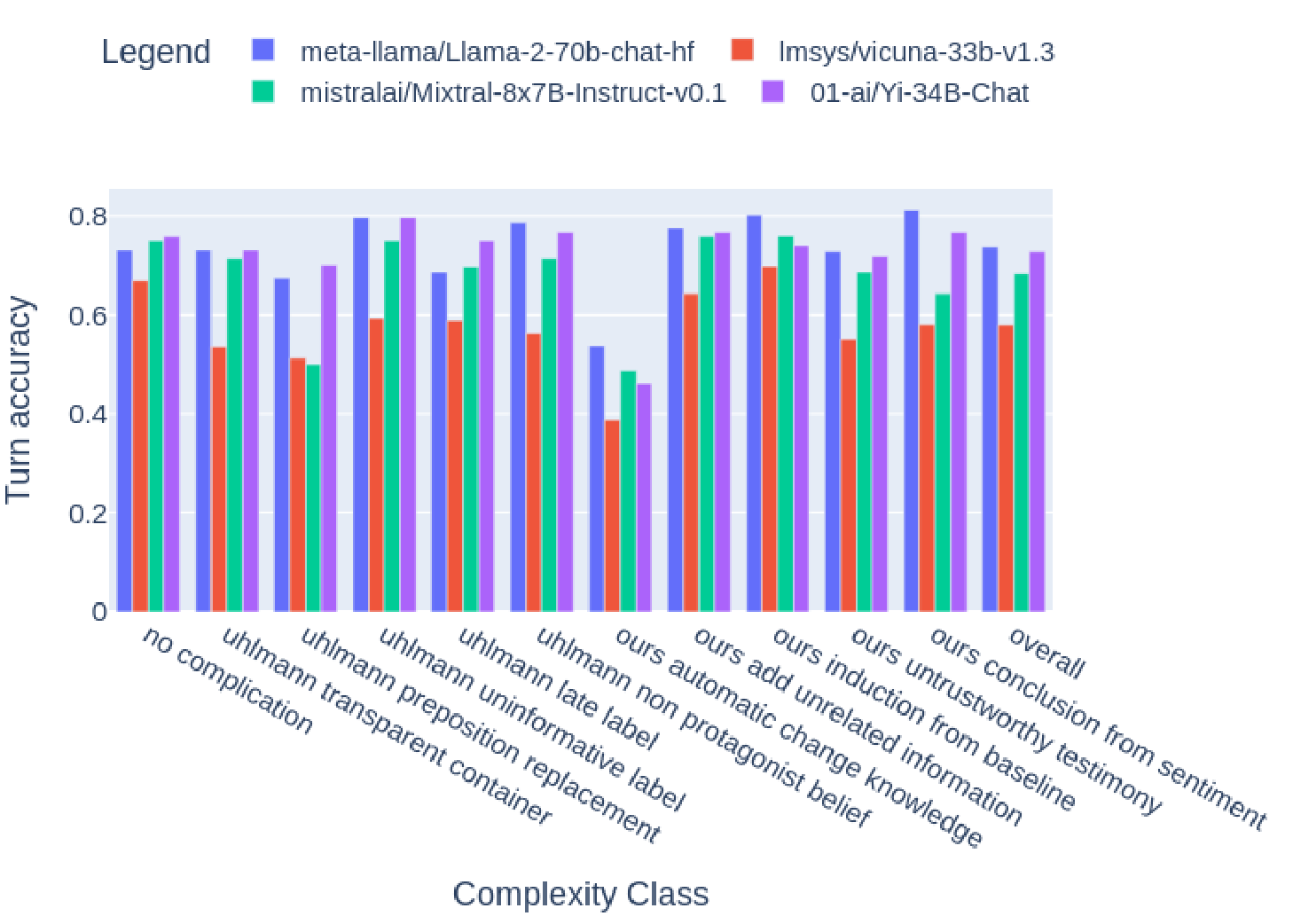}
\end{center}
\caption{Overview of turn accuracy of Llama2 (blue), Vicuna (red), Mixtral(green) and Yi(purple) with regards to all the complexity classes and the overall performance}
\label{fig:quadruple_bar_chart}
\end{figure}

Generally we find a better-than-coin-toss performance on all four models tested (figure \ref{fig:quadruple_bar_chart} and table \ref{tab:correctness_rate_by_complexity_class_all_models}) when evaluating on a per scene basis, for which we calculate the rate of correct answers in all answers given, namely the turn accuracy.

\begin{figure}[t]
\begin{center}
\includegraphics[width=\linewidth]{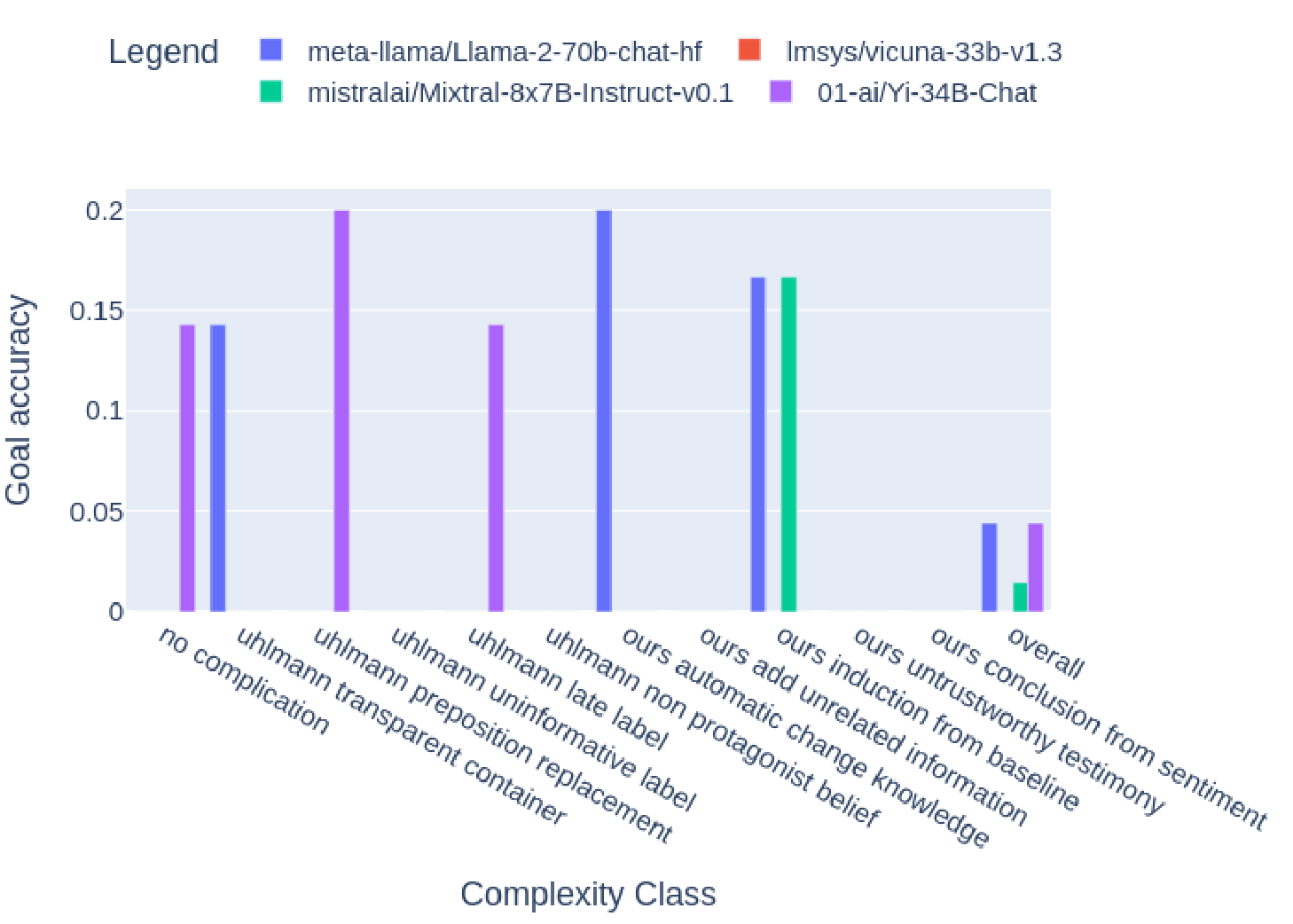}
\end{center}
\caption{Overview of goal accuracy rates of Llama2 (blue), Vicuna (red), Mixtral(green) and Yi(purple) with regards to all the complexity classes and the overall performance}
\label{fig:quadruple_bar_chart_supercorrectness}
\end{figure}

 Whereas when we group all questions asking about the same scene (goal accuracy), we find that most of the time none of the models are able to answer every question about a single scene correctly (figure \ref{fig:quadruple_bar_chart_supercorrectness} and table \ref{tab:supercorrectness_rate_by_complexity_class_all_models}).

Nevertheless we find four categories where we have at least one instance of super-correctness. The overall goal accuracy is above coin toss level ($ 2^{16} \approx 1.5 \times 10^{-5}$). The general tendency is that the LLMs' performance increase with its size in parameters. 

\begin{table}[b]
\begin{tabular}{lllll}
\toprule
Models & Llama-2-70b & vicuna-33b & Mixtral-8x7B & Yi-34B-Chat \\
\midrule
no\ complication & 73.21\% & 66.96\% & 75.00\% & 75.89\% \\
transparent\ container & 73.21\% & 53.57\% & 71.43\% & 73.21\% \\
preposition\ replacement & 67.50\% & 51.25\% & 50.00\% & 70.00\% \\
uninformative\ label & 79.69\% & 59.38\% & 75.00\% & 79.69\% \\
late\ label & 68.75\% & 58.93\% & 69.64\% & 75.00\% \\
non\ protagonist\ belief & 78.57\% & 56.25\% & 71.43\% & 76.79\% \\
automatic\ change\ knowledge & 53.75\% & 38.75\% & 48.75\% & 46.25\% \\
add\ unrelated\ information & 77.68\% & 64.29\% & 75.89\% & 76.79\% \\
induction\ from\ baseline & 80.21\% & 69.79\% & 76.04\% & 73.96\% \\
untrustworthy\ testimony & 72.92\% & 55.21\% & 68.75\% & 71.88\% \\
conclusion\ from\ sentiment & 81.25\% & 58.04\% & 64.29\% & 76.79\% \\
overall & 73.71\% & 58.00\% & 68.47\% & 72.89\% \\
\bottomrule
\end{tabular}
\caption{Turn accuracy by complexity class for all four models.}
\label{tab:correctness_rate_by_complexity_class_all_models}
\end{table}

\begin{table}[b]
\begin{tabular}{lllll}
\toprule
Models & Llama-2-70b & vicuna-33b & Mixtral-8x7B & Yi-34B-Chat \\
\midrule
no\ complication & 0.00\% & 0.00\% & 0.00\% & 14.29\% \\
transparent\ container & 14.29\% & 0.00\% & 0.00\% & 0.00\% \\
preposition\ replacement & 0.00\% & 0.00\% & 0.00\% & 20.00\% \\
uninformative\ label & 0.00\% & 0.00\% & 0.00\% & 0.00\% \\
late\ label & 0.00\% & 0.00\% & 0.00\% & 14.29\% \\
non\ protagonist\ belief & 0.00\% & 0.00\% & 0.00\% & 0.00\% \\
automatic\ change\ knowledge & 20.00\% & 0.00\% & 0.00\% & 0.00\% \\
add\ unrelated\ information & 0.00\% & 0.00\% & 0.00\% & 0.00\% \\
induction\ from\ baseline & 16.67\% & 0.00\% & 16.67\% & 0.00\% \\
untrustworthy\ testimony & 0.00\% & 0.00\% & 0.00\% & 0.00\% \\
conclusion\ from\ sentiment & 0.00\% & 0.00\% & 0.00\% & 0.00\% \\
overall & 4.41\% & 0.00\% & 1.47\% & 4.41\% \\
\bottomrule
\end{tabular}
\caption{Goal accuracy by complexity class for all four models.}
\label{tab:supercorrectness_rate_by_complexity_class_all_models}
\end{table}

Let us now delve deeper into the results on a per model basis. 

\subsection{Llama-2-70-b-chat-hf}
The largest of the Llama 2 models exhibits the overall best performance on our dataset. 
Across all complexity classes of tasks it hits the correct answer approximately 70\% of the time (figure \ref{fig:quadruple_bar_chart}). With exception of the class "automatic change knowledge" which is answered correctly only 53.75\% of the time. The overall turn accuracy is 73.71\%. The best performance was exhibited in "conclusion from sentiment" with 81.25\%. When examining the performance with regards to goal accuracy, we find that the model does not once achieve it in most complexity classes, including the "no complication" class. Interestingly the class "automatic state change", which performed poorly in turn accuracy, is the best performing here with 20\% goal accuracy, followed by "induction from baseline" with a rate of 16.67\% and "transparent container" with 14.29\% (figure \ref{fig:quadruple_bar_chart_supercorrectness}).



\subsection{Vicuna-33b-v1.3}
Being a relative of Llama 2, but with approximately half the number of parameters we find Vicuna-33b-v1.3 to perform  worse. The overall turn accuracy is 58.00\%. "No complication" tasks are solved in 74.00\% of the cases. The performance of the complexity classes ranges from the again worst performing class "automatic state change"  with 38.75\% 
to "Induction from baseline" with 69.79\% (figure \ref{fig:quadruple_bar_chart}). Taking a look at goal accuracy Vicuna has the lowest performance of all models evaluated. For none of the sceneries Vicuna was able to answer all questions correctly (figure \ref{fig:quadruple_bar_chart_supercorrectness}).


\subsection{Mixtral-8x7B-Instruct-v0.1}
As a MoE model Mixtral is similarly sized as Llama-2-70b. The model's performance (figure \ref{fig:quadruple_bar_chart}) on our dataset is third best with an overall turn accuracy of 68.47\% and a "no complication" turn accuracy of 75.00\%. With regards to complexity classes yet again "automatic change knowledge" seems to be the most challenging with a turn accuracy of 48.75\%. The most consistently correctly solved class is "induction from baseline" with 76.04\%. The only instance of goal accuracy we find in the class "induction from baseline" in 16.67\% of the cases. This results in an overall goal accuracy of 1.47\% (figure \ref{fig:quadruple_bar_chart_supercorrectness}).



\subsection{Yi-34B-Chat}
The final model we evaluate on our dataset is Yi-34B-Chat. With regards to overall turn accuracy being 72.89\% it is the second best performing model tested. As shown in figure \ref{fig:quadruple_bar_chart} "no complication" turn accuracy is 75.89\%. The worst performing complexity class is as with the previous models "automatic change knowledge" with 46.25\% turn accuracy. The best performing is "uninformative label" with 79.69\%. Yi achieves goal accuracy in two classes with each instance 14.29\% frequency resulting in an overall goal accuracy of 4.41\% (figure \ref{fig:quadruple_bar_chart_supercorrectness}). In this regard Llama 2 and Yi are on the same level.


\subsection{Dataset Baseline}
To establish a baseline of comparison for our results we not only calculate the expected results if each answer was generated by a coin toss. We also evaluated the four LLMs on the dataset used in \citep{kosinskiTheoryMindMight2023}. All four models perform slightly better on our dataset compared to the former dataset. The highest performance difference is found for the Yi-34B-Chat model with  a difference of almost 10\%.
For the goal accuracy only the Yi-34B-Chat model has a success rate of 14\% on our dataset. The other models have a goal accuracy of 0\%. In comparison, both Llama-2-70B and Mixtral-8x7B show some successful results on the baseline dataset.

\section{Analysis}
The overall performance of the evaluated LLMs is significantly better than the expected baseline for the turn accuracy of 50\%, assuming a coin toss selecting the answer among the two plausible objects or positions mentioned in the stories. The number of cases where the LLM answer does not match one of the two plausible objects is negligible.
The turn accuracy suggests, that the tested LLMs perform slightly better on our dataset compared to the dataset used by \citet{kosinskiTheoryMindMight2023}, though the difference is marginal.
However to assess the models "understanding" of a scene the measure of goal accuracy is preferable as it requires all scenes relating to the same scenery to be answered correctly. The results reported by \citet{kosinskiTheoryMindMight2023} also use this measure.
Barely any scene can be completely solved, thus the goal accuracy is consistently low for any complexity class. This implies that the models learned some linguistic patterns that make it possible to solve more than half the actual questions asked, but possess no robust Theory of Mind. Due to the low rate of goal accuracy, which for most complexity classes is zero, it is also hard to tell differences between the difficulty of complexity classes with regards to the models performances.
Nevertheless it should be noted that the coin toss baseline for goal accuracy, that is answering all 16 questions asked about a single scenery correctly, is estimated as $\frac{1}{2}^{16} ~= 1.53 \cdot 10^{-5}$, which Llama2-70B, Mixtral-8x7B and Yi-34B supersede with an overall goal accuracy between 1.47\% and 4.41\% and a per complexity class goal accuracy up to 4.41\%.
As we know from the previous study consisting of only "no complication" tasks, larger LLMs (late GPT-3.5 and GPT-4) can actually achieve higher rates in goal accuracy that come close to the performance of a 7 year old child. Thus on those larger models we might see more interesting patterns with regards to the challenge posed by the different complexity classes of tasks. Furthermore we note that in this study even tasks with difficult complications sometimes are solved "goal accurately". This is why we suspect that the complications suggested in \citet{ullmanLargeLanguageModels2023} might not by their nature be unsolvable for LLMs, but might get solvable by more advanced models in the future.

Even though the models exhibit no consistent ToM abilities they still answer many questions correctly. This is why we can still attribute different degrees of challenge to the complexity classes. While we see some variation in most complexity classes the impressive drop in performance for "automatic change knowledge" is evident. We think this might be due to the transfer nature of the task involving several steps of thought. First the LLM needs to recognize that the protagonist holds a specific assumption about the dynamics (automatic changes) in the surrounding world. Next it needs to recognize that this mental model will be applied by the protagonist. Finally the model needs to compute the prediction of that mental model of the protagonist with regards to the story. One is likely to produce details in this kind of task that the LLM has not encountered during training. This complexity class might be most interesting to try out using Chain-of-Thought-Prompting in order to alleviate those challenges.

An interesting phenomenon can be observed with the two complexity classes of "transparent container" and "preposition replacement". Both categories of tasks basically deal with the property whether an object that is somehow placed inside or on top of a container is visible or invisible to the protagonist.
On the one hand, given a transparent container task Vicuna shows a drop in turn accuracy. While Mixtral seems to recognize the transparent property of the container and consequence that the protagonist can see what is inside. Thus Mixtral seems to be more capable in dealing with such details. On the other hand when administered the similar complication of a replaced preposition, let's say an object is not placed inside of a container, but rather on top of it such that the protagonist can see it, both, Vicuna and Mixtral, exhibit a sharp decline in turn accuracy (\ref{fig:quadruple_bar_chart}). The decline in Mixtral's ability to answer correctly is intriguing. It may be an artifact of poor spatial reasoning capabilities of LLM - which is not required for the transparent container tasks - , which is more pronounced with "smaller" models. As an, albeit rather large model in total, Mixtral is in fact a mixture of "smaller" experts, hence the label of "8x7B" in the model's name. We suppose that the underperformance in spatial reasoning of smaller models can't be alleviated by the Mixture of Experts approach.

Considering the turn accuracy as well as the goal accuracy the performance of Yi-34B is very close to the performance of Llama2-70B, even though it has almost twice the number of model parameters. The similar sized Vicuna-33B has significantly lower performance. This indicates that the model size alone does not tell much about ToM capabilities.

\section{Discussion and Future Work}
The low goal accuracy across all tested models makes it hard to make meaningful conclusions except that the models exhibit only very limited ToM capabilities and that the "automatic state change" seems to be the most challenging complexity class. 
As mentioned above employing Chain-of-Thought prompting seems especially promising for this class. 

Our dataset can be helpful for further investigation into ToM capabilities and their weak-spots.
However since \citet{kosinskiTheoryMindMight2023} reported better results for their no complications tasks, using larger models of the GPT family, the supposed emergence of ToM capabilities might only occur in models larger than the models we tested and should be evaluated in those. Further insights might be gained when higher goal accuracy are reached and a more fine-grained differentiation between the performance of such LLMs on our data and different complexity classes can be drawn. Therefore repeating our experiments with larger models and models of the GPT family seems promising.

The training of LLMs on data scraped from the internet poses challenges to benchmarks, since LLMs might have been trained on this contaminated data. This introduces the risk of overestimating the models capabilities. Our new hand-written dataset might therefore be valuable for benchmarking LLMs that were trained after the public release of the data from \citet{kosinskiTheoryMindMight2023}.

Also further experimenting with other prompting approaches might be promising to get deeper insights into the LLMs ToM capabilities.

\section{Conclusion}
In this paper we contribute to the ongoing discussion about emergent ToM capabilities in LLMs by creating a new ToM benchmark dataset consisting of 1088 scenes grouped into ten complexity classes and one "non complexity class" addressing the "data draught". Building on the work of \citet{kosinskiTheoryMindMight2023}, we provide a dataset to systematically evaluate the complications proposed by \citet{ullmanLargeLanguageModels2023} and expand on it by introducing new complexity classes. In contrast to "ToMBench" \citep{chenToMBenchBenchmarkingTheory2024} it provides extensibility as it already extracts the answers from elaborate LLM outputs and can be extended to longer CoT outputs in the future. We use our dataset to evaluate the ToM capabilities of four SotA LLMs and find that none of them show robust ToM capabilities, as observed by \citet{kosinskiTheoryMindMight2023}, although they answer some subtasks correctly. In this regard our results align with those of \cite{shapiraCleverHansNeural2023}, who use a similar study design, as well as FANToM \citet{kimFANToMBenchmarkStresstesting2023}, which makes use of dynamic social interactions. Thus, in the realm of ToM, they might be considered stochastic parrots after all \citep{benderDangersStochasticParrots2021a}. The overall low goal accuracy does not allow meaningful conclusions about the impact of the complications on the goal accuracy, which was described to be significant by \citet{ullmanLargeLanguageModels2023}. This might be a result of our selection of tested models, which have a lower number of parameters than those evaluated by \citet{kosinskiTheoryMindMight2023} and might thus not have the same capabilities. The explanatory power of our results about the hardness of different complications remains limited, but the low performance on the "automatic change knowledge" class across all models is intriguing. Due to the nature of these tasks CoT prompting approaches might be promising to improve the ToM performance in certain cases. As we propose very challenging tasks in order to gain further insights into the robustness of ToM pertaining to our complexity classes further evaluation on even larger models is necessary and we encourage every researcher to use our dataset to improve on our baselines.

\section*{Acknowledgments}
The authors gratefully acknowledge the granted access to the Bender cluster hosted by the University of Bonn.

\bibliography{colm2024_conference}
\bibliographystyle{colm2024_conference}

\appendix
\section{Appendix}

\subsection{Complexity Class Examples}
\label{appendix:complexity_class_examples}

\paragraph{automatic change knowledge}

Example:
\texttt{"Charlie buys a non-transparent box of green mangos. There are no red mangos in the box. 
Green mangos ripen, change the color and become red mangos after a few days. Charlie does not know this.[...]"}

\paragraph{add unrelated information}

Example:
\texttt{"Aya finds a non-transparent bottle.  Next to it lie boxes of candy, a bag of popcorn and several unidentifiable objects. There's also a spilled box of truffles. The bottle is decorated with flowery motives. It has a nice brownish tone as base color. It is made from clay and seems to be really ancient. She considers donating it to a museum. She has never seen the bottle before and does not know what is inside.[...]"}

\paragraph{induction from baseline}

Example:
\texttt{"Esther finds a non-transparent paper bag in her parents kitchen cabinet. She does not know what is inside the paper bag. However she then thinks to herself that the bag looks exactly like the paper bags in which they sell sweets in the local supermarket and use for nothing else.[...]"}

\paragraph{untrustworthy testimony}

Example:
\texttt{"[...]Her sister comes by and tells her that the bottle is filled with sweet sauce and not spicy sauce. She knows that her sister loves to play tricks on her and usually the opposite of what she says is true.[...]"}

\paragraph{conclusion from sentiment}

Example:
\texttt{"[...]She reads the gift card. The text says that the present is from her parents and that they are sure she will like the present. Previously she mentioned to her parents that she really enjoys licorice and no other sweets or cookies[...]"}

\subsection{Full Prompt}
\label{appendix:full_prompt}

\lstset{
  basicstyle=\footnotesize,
  breaklines=true,
  breakatwhitespace=true,
  columns=flexible,
  numbers=left, 
  numberstyle=\tiny\color{gray}, 
  keywordstyle=\color{blue}, 
  commentstyle=\color{green}, 
}

\begin{lstlisting}
### YOUR TASK ###
Below you will find a text describing a scene with one or more agents and one or more objects. Objects may have properties, for example they may be inside of one another or they may have a location. Those properties can change throughout the text.

Your task is to tell me for the describing text, where the agents in the text think the object is or what the agent thinks is inside of the object after the last sentence of the text. In other words you tell me what the agent believes to be the property of an object. Valid answers are the object containing the other object or the location of the object or "unknown", if the agent does not know about the location of the object. Only tell me the final information about what the agent thinks after the events of the full text.

You must present this information in JSON-format, where the top level only has the entry "final", as you are required to report the final believes of the agents after the events of the full text.

On the lower level this single entry has the fields for all asked for objects from the question, which contain the agent's belief about that object's property. For example they might have an assumption about it's location or another object containing it. 

You will also be penalized if you display the actual position or objects unless what the agents thinks conicides with the true world state described in the text, because you are required to display what the agents thinks! 

Notice that unless explicitly mentioned agents do not automatically read or see objects or labels. 

I'm going to tip $100 for a better solution!


#### EXAMPLE ####

Text:
There is a non-tranparent box in the kitchen.
Sam is also in the kitchen and can see the box.
The Box is labelled "bananas".
Sam reads the label.

Question:
What does Sam think is inside the box after each sentence?


Expected answer as JSON:
===BEGIN JSON===
{
    "final": {
        "box": "bananas"
    }
}
===END JSON===

#### END OF EXAMPLE ####
\end{lstlisting}



\end{document}